\documentclass{article}

\usepackage{arxiv}

\usepackage[utf8]{inputenc}
\usepackage[T1]{fontenc}
\usepackage{graphicx}
\usepackage{amsmath,amssymb}
\usepackage[round]{natbib}
\usepackage{url}
\usepackage{hyperref}
\usepackage{microtype}

\title{When Everyone Can Build a Model: Mechanistic Inference in the Age of AI}
\renewcommand{\shorttitle}{When Everyone Can Build a Model}

\author{Marvin van Aalst\,$^{1,*}$, Adèle H. Ribeiro\,$^{2,3,\ddagger}$ and Anna Matuszy\'nska\,$^{1,3,\ddagger}$ \\[1ex]
\normalfont $^{1}$Computational Life Science, Faculty of Mathematics and Natural Sciences, \\
\normalfont RWTH Aachen University, Germany \\
\normalfont $^{2}$Causal and Explainable Artificial Intelligence in Life Sciences, \\
\normalfont Faculty of Mathematics and Natural Sciences, RWTH Aachen University, Germany \\
\normalfont $^{3}$Center for Computational Life Sciences, RWTH Aachen University, Germany \\[1ex]
\normalfont $^{*}$Corresponding author: \texttt{marvin.van.aalst@rwth-aachen.de} \\
\normalfont $^{\ddagger}$Equal contribution}

\begin{document}

\maketitle

\begin{abstract}
Modern deep learning and AI-assisted code generation are lowering the technical barrier to building mathematical models of biological systems that reproduce complex observations.
This lets researchers with domain expertise build semi-mechanistic, hybrid models
constrained by prior knowledge and augmented with learned structure
without extensive training in modelling.
However, easier construction does not make interpretation easier: predictive success of such so-called grey-box models does not establish that an inferred mechanism
corresponds to the one that generated the data,
and
the learned component can absorb errors in the mechanistic part.
Many biological modelling problems are underdetermined, so several, possibly infinitely many, models may reproduce the same data equally well.
This ambiguity enters at two modelling choices:
which variables can affect
others, and which response function describes the underlying mechanism.
For the first choice, causal inference provides a formal framework stating
which causal relationships and intervention effects are identifiable from the data and available knowledge.
For the second, ensemble analyses quantify the uncertainty and the range of behaviours still allowed by models that fit equally well.
Together, they turn a single good fit into an explicit account of what is supported, what remains open, and which experiment would decide.
\end{abstract}


\section{Introduction}

Building a mathematical model that reproduces the dynamics of a biological system once required expertise in both biology and mathematical modelling, but this barrier is eroding \citep{martins_dos_santos_systems_2024,stalidzans_mechanistic_2020}.
Grey-box models mix structure fixed by prior knowledge with learned structure \citep{psichogios1992hybrid,thompson1994modeling}, so they can fit data that a purely mechanistic model cannot.
Examples include universal differential equations (UDEs, \citealp{rackauckas2020universal}) and physics- or biology-informed neural networks \citep{raissi2019physics,lagergren2020biologically}.
AI code-generation tools now put such models within reach of domain experts, for whom they are a genuine opportunity.
A close fit, however, invites a mechanistic reading of the model's structure and parameters, and in biomedicine this reading is where the risk lies.
The model's interactions and fitted parameters
can be read as a disease mechanism, a drug target, or an explanation for treatment non-response, and a wrong reading points discovery at the wrong target.
The problem is not new: expert-built models can also fit well while representing the wrong mechanism.
What has changed is that a trained modeller,
who would ask whether alternative structures explain the data equally well, may no longer be involved, and
the new tooling does not supply that skill.

A good fit is weak evidence for a mechanism \citep{oreskes1994verification}, because many biological modelling problems are underdetermined:
many models reproduce the same data equally well.
A bad fit, by contrast, is strong evidence against it \citep{popper1959logic}, although only together with its auxiliary assumptions \citep{duhem1954aim,quine1951dogmas}.
We refer to the underlying property as model distinguishability:
whether the data
provide evidence for one model structure over its alternatives.
More samples of the same kind estimate the observed behaviour more precisely, but cannot break a tie between models that predict it identically.
Systems theory and hydrology describe closely related problems as equifinality \citep{bertalanffy1968general,beven2006manifesto}, machine learning as the Rashomon effect or predictive multiplicity \citep{breiman2001statistical,marx2020predictive}, causal modelling as Markov equivalence \citep{vermapearl1990equivalence}, and philosophy as underdetermination \citep{quine1951dogmas}.

How much this ambiguity matters depends on the question: a prediction within the observed regime can be checked on held-out observational data, whereas a mechanism or an intervention effect cannot. A prediction question asks what to expect for $Y$ after observing $X = x$, that is, $P(Y \mid X = x)$, such as the blood pressure $Y$ of patients who take drug $x$.
A black-box model can legitimately answer this question from observational data, provided new data come from the same distribution as the training data.
A causal question asks what to expect for $Y$ if $X$ is set to $x$ by an intervention, that is, $P(Y \mid \mathrm{do}(X = x))$, such as the blood pressure of patients if they are given drug $x$.
Answering this question requires appropriate causal inference methods \citep{pearl2009causality}.

\begin{figure}
\begin{center}
\includegraphics[width=\textwidth]{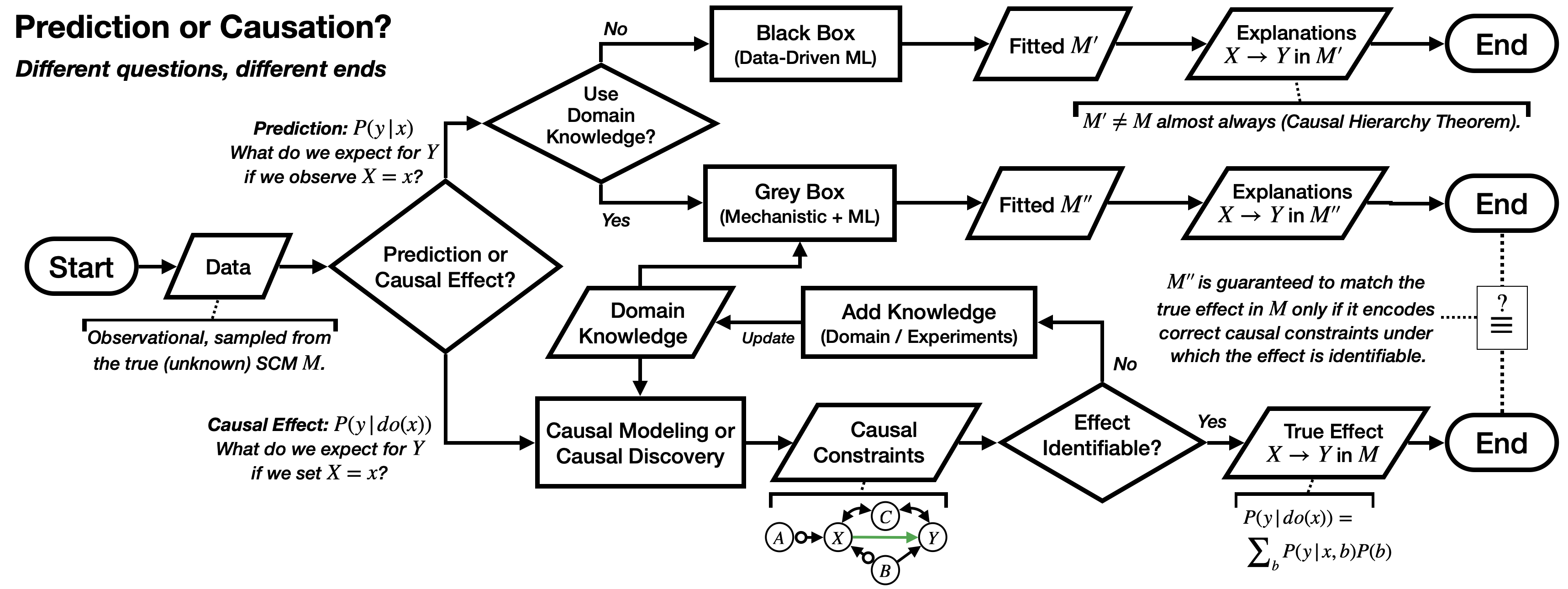}
\end{center}
\caption{Prediction or causation? Different questions, different ends.
Starting from observational data sampled from an unknown structural causal model (SCM) $M$, the question decides the route.
A prediction question asks for $P(y \mid x)$, short for $P(Y \mid X = x)$, and is answered by fitting a black-box model or, if domain knowledge is used, a grey-box model.
Observational data alone do not determine how almost any $M$ responds to interventions, whatever the model class (Causal Hierarchy Theorem, \citealp{bareinboim2022hierarchy}), so the fitted black-box model $M'$ generally differs from $M$ under intervention.
The fitted grey-box model $M''$ is guaranteed to match the true effect only if its domain knowledge encodes correct causal constraints under which that effect is identifiable. The symbol $\stackrel{?}{\equiv}$ between the two lower ends marks this comparison.
A causal question asks for $P(y \mid \mathrm{do}(x))$, short for $P(Y \mid \mathrm{do}(X = x))$.
Causal modelling or causal discovery yields causal constraints on the observed variables, encoded, for example, in a partial ancestral graph that represents the causal structures compatible with the data and domain knowledge, while accounting for latent confounding.
If the effect is identifiable from these constraints, an identification formula expresses it in terms of the available data (here, adjustment for $B$).
If not, further domain knowledge or experiments can refine the constraints.}\label{fig:flowchart}
\end{figure}

Domain experts often build models to understand mechanisms, which is a causal question and requires a different route from prediction (Figure~\ref{fig:flowchart}). In mechanistic or grey-box models, ambiguity arises from two choices: which variables can affect which others, and which response functions describe how variables depend on their inputs. The causal route does not aim to recover the full structural causal model. Instead, it identifies the causal effects that are determined by the causal structure and the observational distribution, without requiring the response functions to be known. This distinction matters because the full structural causal model is almost never identifiable from observational data alone \citep{bareinboim2022hierarchy}.

\section{Which Variables Affect Which}

Which variables affect which is a causal claim: it states which variables would change if another were set by an intervention, not merely which are associated.
In a mechanistic model, the wiring is the set of variables entering the equation for each variable, that is, the variables the model treats as its direct causes.
For a learned term, the wiring is which variables it takes as inputs and which variable it acts on.
Several causal models can induce the same observational distribution and are therefore observationally indistinguishable, even if they differ in wiring or causal structure, response functions, or unobserved influences \citep{pearl2009causality}. Unobserved common causes and selection bias, for example due to recruitment criteria or survival to measurement, can further increase the number of compatible models.

A learned term fitted to an inappropriate set of inputs can reproduce the observed data as well as, or even better than, one fitted to a valid adjustment set for the intervention of interest, yet predict the wrong intervention effect. For predicting $Y$, the model's wiring need not match the true data-generating mechanism because prediction can exploit any association with $Y$, whether causal or not. Consequently, a good fit does not establish that the model correctly represents how its inputs causally influence $Y$.

Correctly estimating an input's causal effect requires that the effect be identifiable from the data and available domain knowledge. The corresponding identification formula specifies how to account for confounding while avoiding the introduction of other sources of bias.

\begin{figure}
\begin{center}
\includegraphics[width=\textwidth]{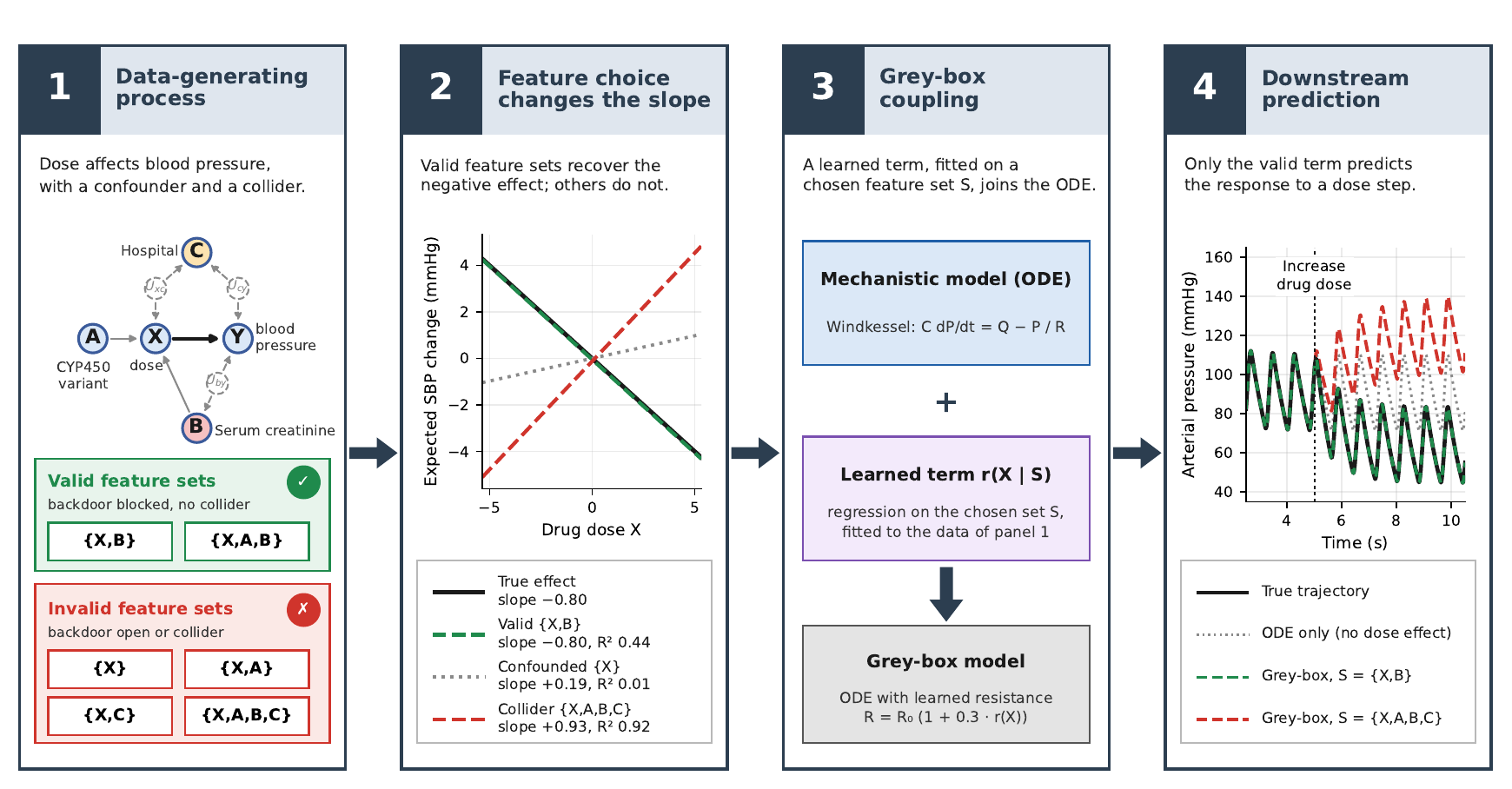}
\end{center}
\caption{Feature choice decides whether the learned term of a grey-box model carries the true causal effect.
All data are simulated from the structural causal model in panel 1, and the coupled mechanistic model is an illustrative Windkessel model.
\textbf{(1)} Drug dose $X$ affects blood pressure $Y$, while dashed circles mark unobserved variables.
$A$ is an instrument, which affects $Y$ only through $X$; $B$ is an observed cause of $X$ that shares an unobserved cause with $Y$; and $C$ is a collider of unobserved causes of $X$ and $Y$.
Feature sets that block the backdoor path through $B$, a non-causal route from $X$ to $Y$, and exclude the collider are valid; all others are invalid.
\textbf{(2)} Expected change in blood pressure as a function of dose, from linear regressions on different feature sets $S$.
The valid set $\{X,B\}$ recovers the true effect (black), whereas the confounded set $\{X\}$ and the collider set $\{X,A,B,C\}$ give positive slopes.
The in-sample fit ($R^2$) is highest for the collider set, so goodness of fit does not reveal which set is valid.
\textbf{(3)} The regression on $S$ yields a learned term $r(X \mid S)$, which scales the vascular resistance $R$ of the Windkessel model and forms the grey-box model.
\textbf{(4)} After a step increase in dose, the grey-box model with the valid set follows the true trajectory.
The model with the collider set predicts a rise in pressure instead of a fall.}\label{fig:feature-choice}
\end{figure}

Figure~\ref{fig:feature-choice} illustrates this distinction using simulated data, in which a learned dose effect scales vascular resistance in a model of arterial pressure. The dose effect is identified by adjusting for $B$, which blocks confounding, but not for $C$, a collider of unobserved causes of dose and pressure. Conditioning on $C$ opens a non-causal path. Yet a feature set that includes $C$ explains far more variance than the valid adjustment set $\{B\}$. Selecting model inputs based on goodness of fit can therefore favour a causally incorrect specification.

Two classical sources of bias help explain this discrepancy. Confounding occurs when a common cause influences both an input and an outcome. Omitting this cause can distort an estimated association and even reverse its sign, as in Simpson's paradox \citep{pearl2009causality,bonovas2023simpson}. Collider bias arises when an analysis conditions on a common effect of two variables, inducing an association between them even when they are otherwise causally unrelated \citep{elwertwinship2014collider}. Selection into a sample can induce the same type of bias when selection depends on a common effect of variables related to the input and outcome.
Neither problem is necessarily apparent from model fit. Instead, it can lead to incorrect estimates of intervention effects, even when predictions under the observational distribution are accurate.

Causal discovery recovers the wiring only up to an equivalence class, but it states which parts of the wiring the data leave open.
Under latent confounding and selection bias, the class of structures compatible with the observed conditional independencies is represented by a partial ancestral graph (PAG) \citep{richardson2002ancestral,zhang2008completeness}.
The Fast Causal Inference (FCI) algorithm is sound and complete for this task, so in the large-sample limit circle marks in the PAG reflect structural ambiguity that observational data alone cannot resolve without further assumptions \citep{zhang2008completeness}.
These guarantees rely on faithfulness, which rules out cancelling effects, and, in finite samples, on dependences strong enough to detect. Violations of either assumption can lead to errors in independence tests that propagate through FCI's orientation rules, yielding an incorrect or even invalid PAG without warning.
Data-Compatible FCI (dcFCI) addresses this by assessing how compatible each candidate PAG is with the data and returning a ranked set of candidates instead of a single one \citep{ribeiro2026dcfci}.

Within these limits,
causal discovery turns a vague worry about the wrong wiring into an explicit list of open alternatives.
An intervention effect can still be identifiable when it has the same value, given the observed distribution, in every model of the equivalence class. The generalized adjustment criterion determines when the effect can be obtained by covariate adjustment \citep{perkovic2018adjustment}, and complete identification algorithms IDP and CIDP decide identifiability in general, for marginal and conditional effects, respectively \citep{jaber2022causal}.
\citet{fehse2026gutbrain} applied this pipeline end to end, discovering and estimating identifiable effects of gut taxa on major depressive disorder solely from observational data.

Domain knowledge and experiments narrow the equivalence class, and this is where an expert counts most, and where the expert's own assumptions enter.
Knowledge helps when it can be written as a constraint on the structure, as anchorFCI does with variables known not to be caused by the others, such as genetic variants \citep{ribeiro2024anchorfci}, or when an expert is queried about the relations the data leave most uncertain \citep{dasilva2026agfn}.
Observational and interventional data from different contexts can also be analysed jointly \citep{mooij2020jci,li2023multienv}, which resolves ambiguity wherever the interventions distinguish competing structures.
Incorporated knowledge must, however, be consistent with the system under study, since a wrong constraint can exclude the true structure without the data showing the mistake.

\section{Which Response Function}

A causal structure qualitatively specifies which variables affect which; it does not quantify or describe their response functions.
Within physiological ranges, simplified rate laws can reproduce much of the dynamics of a detailed kinetic model, and how well depends on the operating regime of each reaction \citep{du2016ratelaw}.
If the data cover only a narrow range, such as concentrations far below saturation, the nonlinearity that distinguishes the rate laws is never excited.
Outside that range, for instance under a high drug dose, their predictions can differ sharply.
A learned term represents the limiting case: the network itself is the response function, drawn from a very flexible family, so many different functions can fit the same data equally well.

Even with the response function fixed, its parameters can be structurally or practically non-identifiable, so tight predictions can coexist with loosely constrained parameter values.
A parameter is structurally non-identifiable when even noise-free data on the measured outputs cannot determine it, and practically non-identifiable when the available data do not \citep{raue2009identifiability}.
\citet{gutenkunst2007sloppy} found parameter sensitivities spread over many orders of magnitude in all seventeen published systems-biology models they surveyed, built by domain experts before neural differential equations.
The authors of Bayesian calibration warn that calibrated parameters are best-fitting values, not estimates of the true physical ones \citep{kennedy2002bayesian}.
What parameter non-identifiability undermines is thus not the prediction within the observed regime but the physical reading of the parameter values.

A learned term fitted jointly with mechanistic response functions can absorb what those equations get wrong, so the fit no longer tests the mechanistic part.
In hybrid models, the learned term can reproduce similar trajectories for different, including incorrect, values of the mechanistic parameters \citep{giampiccolo2024hybrid}.
When the learned and mechanistic parts act on the same variable, even noise-free data may leave the hybrid equivalent to a fully learned model \citep{loman2026structural}.
This matters most when the intervention of interest targets a mechanistic component.
An inhibitor changes one rate constant and a knockout removes a term, so their effects can be predicted only if that component is correctly estimated and its role has not been absorbed.
The architecture also limits what the learned term can represent: one that sees only the current state cannot recover a delayed effect or an unobserved state.

Instead of reporting the single best fit, an ensemble of models that fit equally well shows how far the data constrain the behaviour of interest.
Over parameters, an ensemble collects fits from many starting points, along likelihood profiles, or from a posterior distribution.
Over response functions, it collects alternative rate laws or learned terms trained from different initializations and with different architectures.
Recent work on UDEs uses multi-start fits and likelihood profiles to show which parameters and functions the data determine \citep{loman2025functional,philipps2025ude}.
The quantity to report is the range of the prediction that matters, such as the response to a planned intervention, rather than the parameter values.
A non-identifiable parameter then becomes a reported range instead of a hidden defect.
This is uncertainty quantification in the sense that matters for interpretation: it states what the data leave open, not only what the best model predicts.

An ensemble explores only the structures it was built from, so models that share a wrong wiring agree on the wrong answer.
The logic is one-sided: disagreement among well-fitting models is strong evidence that the data do not determine a prediction, while agreement is only weak evidence that they do.
An ensemble should therefore also vary the wiring wherever causal analysis has left it open, such as the inputs of a learned term or a possible latent state.

\section{From a Fit to a Claim}

Claims should be reported at the level supported by the evidence, distinguishing established knowledge, assumptions, empirical findings, and causal conclusions. Predictive fit and validation support performance claims within tested contexts, but do not establish causal effects or mechanisms. These require causal analysis that incorporates domain knowledge, derives constraints from data, and assesses identifiability and remaining ambiguities. Leakage and batch effects can compromise validation \citep{kaufman2012leakage,leek2010batch}.
Specifying the research question and hypothesized relationships before analysing the data helps distinguish prespecified assumptions from structures inferred post hoc, reducing the risk of presenting exploratory findings as a priori hypotheses (HARKing, \citealp{kerr1998harking}). Credibility standards for computational systems biology similarly emphasize documenting assumptions alongside predictive performance \citep{tatka2023credibility}.

The way forward is iterative:
identify causal models compatible with the evidence, determine which effects are identifiable, and design experiments to distinguish remaining explanations.
Causal analysis narrows the wiring, and ensembles bound the response functions and their parameters.
Experimental design can identify conditions under which competing mechanistic models make different predictions, enabling experiments that discriminate between them
\citep{boxhill1967discrimination,kreutz2009design}.
When interventions are infeasible, natural experiments or independent contexts may provide additional evidence, subject to their assumptions. Domain experts are best placed to design and run such experiments, since they know which perturbations are feasible and which measurements are reliable.

The tools that lowered the barrier to building a model thus leave domain experts with the step that most needs their expertise:
determining which causal explanations remain plausible and which experiments can distinguish between them.

\section*{Conflict of Interest Statement}

The authors declare that the research was conducted in the absence of any commercial or financial relationships that could be construed as a potential conflict of interest.

\section*{Author Contributions}
All authors contributed equally to this work

\section*{Funding}
This research was funded by the Deutsche Forschungsgemeinschaft (DFG, German Research Foundation, FOR 5573/1 to AM), and the Federal Ministry of Research, Technology, and Space (BMFTR) under the CausalAI4Health Project (\#01ZU2503 to AHR).

\bibliographystyle{plainnat}
\bibliography{library}

\end{document}